\pdfoutput=1

\documentclass[11pt]{article}

\usepackage{acl}
\usepackage{times}
\usepackage{latexsym}
\usepackage{tablefootnote}
\usepackage[T1]{fontenc}
\usepackage[utf8]{inputenc}
\usepackage{microtype}
\usepackage{booktabs}
\usepackage{inconsolata}
\usepackage{hyperref}
\usepackage{xurl}
\usepackage{comment}
\usepackage[table]{xcolor}

\usepackage{graphicx}
\usepackage{tabularx}
\usepackage{multirow}
\usepackage{multicol}
\usepackage{multibib}
\usepackage{amsmath, amsfonts, amssymb, amsthm} 
\usepackage{easybmat} 
\usepackage{float}
\usepackage{enumitem}
\usepackage{highlight}
\usepackage[section]{placeins}
\usepackage{authblk}
\usepackage{fdsymbol}
\renewcommand{\opacity}{50}

\newcommand{\sysnamesingle}{\textsc{iDeb}}
\newcommand{\sysname}{\textsc{Mafia}}

\newcommand{\del}[1]{\Delta_{\scriptsize{\textrm{#1}}}}
\newcommand{\delavg}{\del{average}}
\newcommand{\delgnd}{\del{gender}}
\newcommand{\delrac}{\del{race}}
\newcommand{\delrel}{\del{religion}}

\newcommand{\ovr}[1]{\Psi_{\scriptsize{\textrm{#1}}}}
\newcommand{\ovravg}{\ovr{average}}

\newcommand{\sns}[1]{\textrm{\sysnamesingle}_{\scriptsize{\textrm{#1}}}}
\newcommand{\snsgnd}{\sns{gender}}
\newcommand{\snsrac}{\sns{race}}
\newcommand{\snsrel}{\sns{religion}}

\title{\sysname: \underline{M}ulti-\underline{A}dapter \underline{F}used \underline{I}nclusive Langu\underline{A}ge Models\\ 
\normalsize{\textcolor{red}{\textit{This paper has content that might be offensive, or upsetting, however, this cannot be avoided owing to the nature of the work.}}} }
\makeatletter
\renewcommand\AB@affilsepx{\quad\protect\Affilfont}
\makeatother

\let\svthefootnote\thefootnote
\newcommand\freefootnote[1]{%
  \let\thefootnote\relax%
  \footnotetext{#1}%
  \let\thefootnote\svthefootnote%
}
\author[$\dag\clubsuit$]{Prachi Jain}
\author[$\dag\diamondsuit\ddag$]{Ashutosh Sathe}
\author[$\clubsuit$]{Varun Gumma} 
\author[$\heartsuit\ddag$]{\authorcr Kabir Ahuja}
\author[$\clubsuit$]{Sunayana Sitaram}
\affil[$\clubsuit$]{Microsoft Corporation}
\affil[$\diamondsuit$]{Indian Institute of Technology, Bombay\authorcr}
\affil[$\heartsuit$]{University of Washington\protect\\Contact: \texttt{p6.jain@gmail.com}}

\begin{document}
\maketitle

\begin{abstract}


Pretrained Language Models (PLMs) are widely used in NLP for various tasks. Recent studies have identified various biases that such models exhibit and have proposed methods to correct these biases. However, most of the works address a limited set of bias dimensions independently such as gender, race, or religion. Moreover, the methods typically involve finetuning the full model to maintain the performance on the downstream task. In this work, we aim to \textit{modularly} debias a pretrained language model across \textit{multiple} dimensions. Previous works extensively explored debiasing PLMs using limited US-centric counterfactual data augmentation (CDA).
We use structured knowledge and a large generative model to build a diverse CDA across multiple bias dimensions in a semi-automated way. We highlight how existing debiasing methods do not consider interactions between multiple societal biases and propose a debiasing model that exploits the synergy amongst various societal biases and enables multi-bias debiasing simultaneously. An extensive evaluation on multiple tasks and languages demonstrates the efficacy of our approach.
\end{abstract}

\freefootnote{$^\dag$Equal Contribution}
\freefootnote{$^\ddag$Work done when the author was at Microsoft}

\section{Introduction}
Pretrained Language Models (PLMs) are growing in power and prominence across numerous NLP tasks \cite{WANG202351,ahuja-etal-2023-mega}. Their reach has expanded beyond academia, reaching general users through services like code assistance and chatbots \cite{li2023starcoder, kopf2023openassistant}. Despite the extraordinary performance of these models on their respective tasks, several works have identified the harmful social biases picked up by these models as an artifact of their pretraining on web-scale corpus consisting of unmoderated user-generated content \citep[\textit{inter alia}]{manzini-etal-2019-black, webster2020measuring, nadeem-etal-2021-stereoset}.

While most previous works focus on (binary) gender biases, other societal biases, such as race and religion, are less studied in the context of PLMs. Moreover, these biases are often intertwined with each other, creating complex and nuanced forms of discrimination. We define intersectional biases as the biases that arise from the combination of different attributes, such as gender, race, and religion. 
In this work, we focus on building debiasing techniques that can model and mitigate gender (including non-binary), race, religion, profession, and intersectional biases, which are often ignored in previous works. 



The community has developed a gamut of methods to measure and mitigate biases in LLMs \citep{bordia-bowman-2019-identifying, liang-etal-2020-towards, ravfogel-etal-2020-null, webster2020measuring, lauscher-etal-2021-sustainable-modular, DBLP:conf/emnlp/SmithHKPW22, kumar-etal-2023-parameter}. The majority of these methods finetune \textit{all} the parameters of a language model to debias it towards a particular bias dimension such as gender or race, and the escalating size of PLMs can pose computational challenges, particularly for smaller academic labs or enterprises. While some methods \cite{DBLP:journals/tacl/SchickUS21, DBLP:conf/aaai/YangY0LJ23} do not alter a model’s internal representations or its
parameters. Thus, they cannot be used as a bias mitigation strategy for downstream NLU tasks. To this end, we aim to use \textit{adapters} \citep{pmlr-v97-houlsby19a, pfeiffer-etal-2020-adapterhub}, which are small neural network layers inserted in Transformer blocks \cite{vaswani2017attention} of an LLM as a way to effectively debias it towards a certain dimension. We further show that a soft combination of multiple such adapters can be used to exploit the synergy between various bias dimensions and can lead to a fairer and more accurate model on a downstream task.

To train each of the individual debiasing adapters, we make use of the counterfactual data augmentation (CDA) technique. While CDA has been shown to be effective on gender debiasing \citep{zmigrod-etal-2019-counterfactual,dinan-etal-2020-queens,webster2020measuring,barikeri-etal-2021-redditbias,DBLP:conf/emnlp/QianRFSKW22,DBLP:conf/acl/Goldfarb-Tarrant23a}, previous works \citep{meade-etal-2022-empirical, lauscher-etal-2021-sustainable-modular} have relied on a small set of handbuilt (mostly US-centric) counterfactual pairs. As LLM bias is a complex and multifaceted issue, comprehensively addressing it requires considering diverse identities. Hence, we propose a semi-automated method, general purpose to build a comprehensive CDA pair list using Wikidata \cite{denny2014wikidata} and generative models.


Our results indicate that such a general method can be used to train strong debiasing adapters (\sysnamesingle) for multiple dimensions. In particular, we perform experiments on gender, race, religion, and profession. We list our contributions and key findings below:



\begin{enumerate} [label= \textbf{\arabic{enumi}.},ref=Step \arabic{enumi}, wide=0pt]
    \item An inclusive and diverse counterfactual pair dataset\footnote{Unlike previous work, which mainly was US-centric.} for gender, race, religion, and profession bias. Note that, we also take into account non-binary genders. (\S \ref{sec:cda})
    \item \sysnamesingle\ - A more inclusive and improved bias-specific debiasing model, trained on the newly generated diverse and inclusive CDA pairs. (\S \ref{sec:indiv-dba})
    \item \sysname\ - A soft way to combine multiple debiasing adapters on downstream tasks. The model exploits the synergy between various biases to improve fairness as well as performance on the downstream task. (\S \ref{sec:fusion-dba})
    \item We show that \sysname\ can reduce unintended bias on a toxicity classification task on related bias dimensions that are unseen by any of the individual debiasing adapters. (\S \ref{sec:jigsaw})
    \item We observe zero-shot transfer of gains in fairness and performance by debiasing a multilingual PLM on English. We test our models on a new dataset (mBias-STS-B) for measuring fairness across different languages with varying resource availability. (\S\ref{sec:multilingual-stsb})
    \item We release the mBias-STS-B dataset along with the code for future research\footnote{\url{aka.ms/AAoumtu}}.
    \end{enumerate}

\section{Methodology}
In our study, we examine four primary bias dimensions: gender, race (ethnicity), religion, and profession. First, we discuss the method for generating counterfactual (CF) pairs in Section \ref{sec:cda}. Subsequently, we outline the approach to train debiasing adapters (DBAs) for each dimension in Section \ref{sec:indiv-dba}. Lastly, in Section \ref{sec:fusion-dba}, we introduce our strategy for integrating individual DBAs for application on a downstream task.

\subsection{Counterfactual Data Augmentation}\label{sec:cda}
\begin{figure}[t]
    \centering
    \includegraphics[trim={0.25cm 19.75cm 7cm 0.5cm},clip,width=0.5\textwidth]{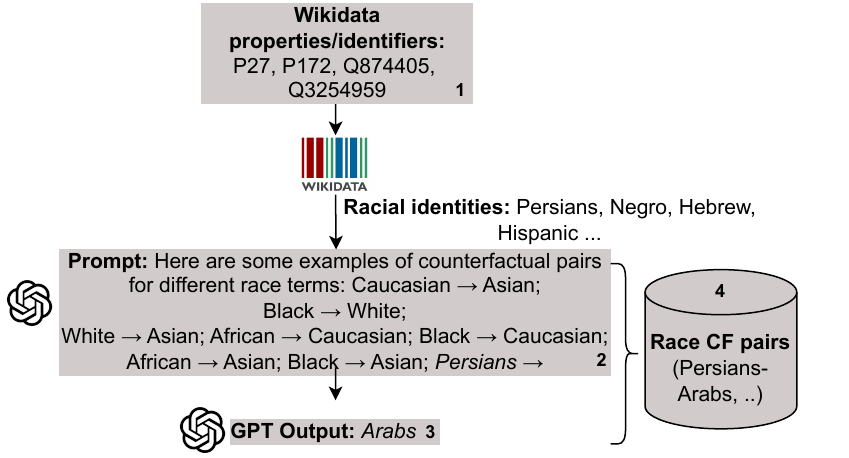}
    \caption{Steps to generate Counterfactual (CF) pairs for racial bias. Note that the technique can be similarly used for other biases.}
    \label{fig:cf-steps}
\end{figure}


\noindent{\bf Counterfactual Data Augmentation (CDA)} is a generic dataset-based debiasing technique \cite{DBLP:conf/nips/KusnerLRS17,DBLP:conf/birthday/LuMWAD20}. 
Given a set of counterfactual (CF) pairs (i.e., $d$ representing the dominant group, e.g., man, and $m$ representing the minority group, e.g., woman) and a training dataset, CDA replaces every instance of $d$ with $m$ and vice-versa ($2$-way CDA) \cite{webster2020measuring} in the training data. The final corpus for debiasing training consists of both the original and counterfactually created sentences. The goal is that such data can balance the effect of pre-existing biases in data and encourage the model to learn fairer representations.\\

\noindent{\bf Generating CF pairs:} Unlike previous methods \citep{lauscher-etal-2021-sustainable-modular, meade-etal-2022-empirical} that rely on handcrafting the CF pairs (mostly US-centric), we propose a semi-automated, generic method to generate CDA pairs. We use a large structured knowledge base as a starting point. Wikidata's \cite{denny2014wikidata} repository of information is rich and diverse, making it an ideal resource for our purpose. We manually identify a list of Wikidata items and properties whose subject or object position has English entities of respective bias type (gender/race/religion/profession) (step 1 in Figure \S\ref{fig:cf-steps}). We refer the readers to Appendix \S\ref{appendix-cf-pairs} for the properties we used for extracting gender, race, religion, and profession terms. 

Using all possible pairs of bias-related words for generating CDA can quickly become intractable, especially when dealing with extensive lists of terms. Additionally, including all pairs may introduce noise in training. It is crucial to exercise caution and thoughtfully curate the pairs to ensure the training process remains effective and reliable. Therefore, we use a generative model\footnote{\texttt{text-davinci-003} \cite{DBLP:conf/nips/Ouyang0JAWMZASR22-dv003}} to build a corpus of CDA from the bias term list. Our prompt has the following structure: \texttt{Here are some examples of counterfactual pairs for different <bias-type> terms: <sample-CDA-pairs>, <bias-term> $\rightarrow$ <output>} (step 2 in \S\ref{fig:cf-steps}). Here, \texttt{<bias-type>} is one of the bias dimensions i.e. gender/race/religion/profession while \texttt{<sample-CDA-pairs>} is a seed set of CDA pairs. We obtain this seed set for gender, race, and religion from \citet{meade-etal-2022-empirical}. For profession, we use the gender seed set.  Finally, we prompt the LLM to produce a suitable counter for a new \texttt{<bias-term>} (step 3 in Figure \S\ref{fig:cf-steps}). 

We find that the generative model can generate a lot of improbable and uninteresting CF pairs during this process. Therefore, to filter out these pairs, we use the Google Book Corpus\footnote{\url{https://api.datamuse.com/}} and retain only those CF pairs where both the entities in the pair appear at least once in a million times in the corpus. For gender, we reduce the threshold to $0.01$. Our final set of CF pairs includes $68$ pairs for gender, $156$ for race, and $86$ for religion. These numbers are notably higher as compared to \citet{meade-etal-2022-empirical} which used $57$, $7$, and $6$ terms for gender, race, and religion respectively.

\begin{figure}[t]
    \centering
    \includegraphics[trim={2cm 14.5cm 7cm 4cm},clip,width=0.5\textwidth]{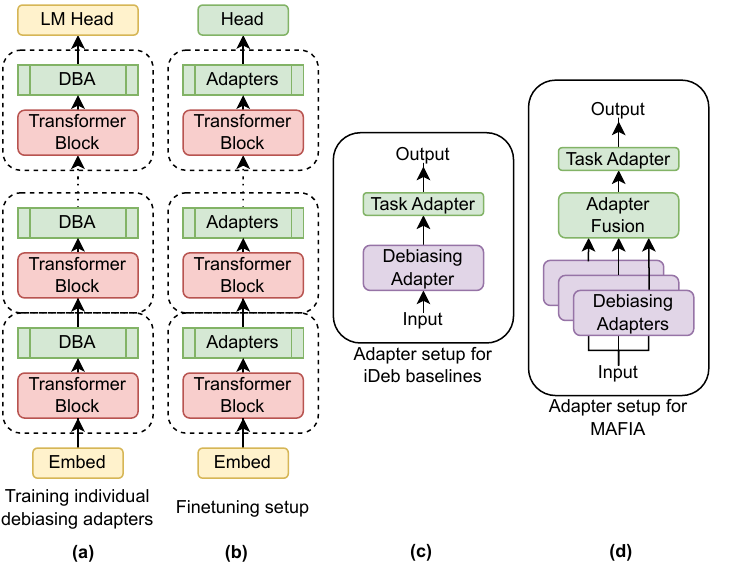}
    \caption{A comprehensive summary of the various training strategies described. Only the components highlighted in green are finetuned in each case.}
    \label{fig:task_adapter}
\end{figure}
\subsection{Training Individual Debiasing Adapters}
\label{sec:indiv-dba}
We adopt the training procedure of \citet{lauscher-etal-2021-sustainable-modular} to train a debiasing adapter (see Figure \ref{fig:task_adapter} (a)). The process involves adding a debiasing adapter to the base LM and is trained with a Masked Language Modeling (MLM) objective \cite{devlin-etal-2019-bert} on our large, inclusive CDA Wikipedia dataset. Note that training the debiasing adapter does \textit{not} introduce task-adapters. 

\citet{lauscher-etal-2021-sustainable-modular} fully finetuned the adapter when using the model on a downstream task (see Figure \ref{fig:task_adapter} (b)). 
However we only train a new task adapter for the end task keeping the remaining model parameters (including the debiasing adapter) frozen (see Figure \ref{fig:task_adapter} (c)). 

\subsection{Combining Multiple Debiasing Adapters}
\label{sec:fusion-dba}
Our final model exploits the synergy of various debiasing adapters to improve performance on respective biases. Therefore, given $k$ debiasing adapters trained \textit{independently} on $k$ bias dimensions, we propose to combine them on a downstream task as shown in Fig. \ref{fig:task_adapter} (d). All the $k$ debiasing adapters are fused via a trainable AdapterFusion \citep{pfeiffer-etal-2021-adapterfusion} layer and stacked with a task-specific adapter to facilitate further intermixing of signals. We refer to such a model with multiple fused adapters as \sysname. We expect \sysname\ to be especially useful in an enterprise setting where product specific teams can easily add (or remove) DBAs for newly identified (or obsolete) bias dimensions to the base model which is often shared across different products. 
\section{Experimental Setup}

\subsection{Evaluation datasets and metrics}
We evaluate \sysname\ on various intrinsic and extrinsic (downstream) bias evaluation benchmarks, and demonstrate its superior debiasing ability over related baselines.

\subsubsection{Intrinsic Evaluation}
We use \textit{Stereoset} and \textit{Crowdsourced Stereoset Pairs} (CrowS-Pairs) to evaluate intrinsic bias in models.

\noindent
\textbf{StereoSet} \cite{nadeem-etal-2021-stereoset} is a large-scale natural English crowdsourced dataset to measure stereotypical biases in four domains: gender, profession, race, and religion. Each StereoSet example consists of a context sentence -- \textit{``Our housekeeper is a $\langle$BLANK$\rangle$.''} And a set of three attributes -- stereotype (Mexican), anti-stereotype (American), and a meaningless option (Banana). We determine which attribute will most likely fill the blank to measure language modeling and stereotypical bias. We use two different measures: (1) {\it Stereotype Score} is the percentage of examples for which a model prefers stereotypical association instead of anti-stereotypical associations. (2) {\it Language modeling score} is the percentage of examples for which a model prefers meaningful associations (either stereotypical or anti-stereotypical) as opposed to meaningless associations. \\
\noindent
\textbf{CrowS-Pairs} \citep{nangia-etal-2020-crows} introduced a crowdsourced benchmark dataset for measuring the degree to which nine types of social bias are present in language models. This work focuses on gender, race (and ethnicity), religion, and professional biases. The dataset consists of stereotypical and anti-stereotypical sentences in a given context similar to StereoSet.
We use {\it Stereotype Score}, the percentage of examples for which a model assigns a higher masked token probability to the stereotypical sentence than the anti-stereotypical sentence. The masked token probability of a sentence is the average probability of unique tokens (w.r.t. counterpart sentence) in the sentence.\\

\noindent
Recent works have raised concerns on the validity of the above two intrinsic evaluation benchmarks' operationalizations of stereotyping \cite{blodgett-etal-2021-stereotyping,DBLP:conf/acl/BlodgettBDW20}. Hence we also evaluate our model on downstream NLP tasks.   

\subsubsection{Extrinsic Evaluation} 
\noindent\textbf{Dataset:} We use STS-B i.e., the Semantic Textual Similarity Benchmark from GLUE \cite{wang-etal-2018-glue} as our downstream task for this evaluation. STS-B requires the model to consider two sentences and output a score between $0$-$5$ indicating how semantically similar the input sentences are. \citet{webster2020measuring} introduced Bias-STS-B which takes a neutral STS template and fills it with a gendered term and a profession term to form two sentences respectively. Original Bias-STS-B used only binary gender terms while in our study we consider $7$ gender identities -- male, female, non-binary, and LGBT. The gender bias evaluation dataset contains $16,980$ such septets (for $7$-way comparison). 
We generate test sets for evaluating race, and religion biases, using the templates released by \citet{DBLP:conf/aaai/DevLPS20}. The sentence pair is built using  
a noun-template -- {\it The $\langle$subject$\rangle$ person $\langle$verb$\rangle$ a/an $\langle$object$\rangle$} and an adjective template -- {\it The $\langle$adjective$\rangle$ person $\langle$verb$\rangle$ a/an $\langle$object$\rangle$}. The $\langle$adjective$\rangle$ is filled with polarised adjectives (e.g., arrogant, brilliant) and the $\langle$subject$\rangle$ is filled with a religion term (e.g., Christian, Hindu, \textit{etc.}) or ethnicity (Black, Caucasian, \textit{etc.}) for generating a religion or race bias evaluation dataset respectively. We produce a total of $688,801$ Race-Bias and $757,680$ Religion-Bias sentence pairs, and we further sub-sample a set of $16,384$ sentence pairs from it for tractable evaluation.
We use $11$ religion terms and $10$ race terms from \citet{DBLP:conf/aaai/DevLPS20} to build the dataset.\\
\noindent
\textbf{Metrics:} On STS-B, we measure the performance by calculating the Pearson correlation \cite{freedman2007statistics} ($\rho$) between model scores and human annotated similarity scores. On Bias-STS-B, we report the \textit{average absolute difference} between scores of individual components. Unlike previous works, we perform a {\it multi-way comparison} instead of a $2$-way comparison. E.g. Bias-STS-B along race component has $k=10$ components (i.e. different races) which means there are 10 sentence pairs (\textit{An African-American kid is playing on the ground} vs \textit{A child is playing on the ground}; \textit{An Indian kid is playing on the ground} vs \textit{A child is playing on the ground} and so on) for which we receive scores $s_1,\dots,s_k$ from the model. Next we calculate average absolute difference as $\Delta = \frac{1}{\binom{k}{2}}\sum_{i=1}^k\sum_{j=i}^k|s_i - s_j|$. Notice that it is trivial to drive $\Delta$ to 0 at the cost of performance on STS-B by producing the same score for every pair. To better account for this tradeoff, we introduce a new metric called {\it``useful fairness''} that lets us compare models on both fairness as well as accuracy axes.
\noindent
We compute ``\textit{useful fairness}'' ($\Psi$) for a particular bias dimension as $\ovr{dim} = \rho \cdot \alpha (1 - \del{dim})$ where $\rho$ is the Pearson score model achieves on original STS-B, and $\alpha (=1)$ is a constant capturing estimated effect of debiasing performance on the final model score, and $\del{dim}$ is the average difference across all components of a particular bias dimension (gender/race/religion) computed on Bias-STS-B as discussed above.

\subsection{Baselines} 
We use BERT, mBERT \cite{devlin-etal-2019-bert}, RoBERTa \cite{liu2019roberta}, XLM-R \cite{conneau-etal-2020-unsupervised} as our base LMs. To validate the effectiveness of \sysname, we primarily compare it against the base LM as well as $\sns{bias}$\ (debiased for respective biases). Fig. \ref{fig:task_adapter} shows our general finetuning setup along with Adapter setups for various baselines.


On \textit{gender}, we additionally consider using CF pairs from \citet{lauscher-etal-2021-sustainable-modular} to train a DBA. We also compare with an ``AdapterDrop'' approach \cite{ruckle-etal-2021-adapterdrop}, an adapter-based dropout regularization method since previous work by \citet{webster2020measuring} showed that dropout helps model debiasing. We call this model {\it $\langle$baseLM$\rangle$+\textrm{AD}}. The model architecture is similar to base LM + task-adapter in Fig. \ref{fig:task_adapter} except the task-adapter is an ``AdapterDrop'' enabled adapter.
Another baseline we compare with is a single DBA model trained on the concatenation of all CDA data used for four bias dimensions, denoted by ${\sns{all}}$.

\subsection{Hyperparameters}
All Debiasing Adapters (DBAs) and task-adapters use Pfeiffer architecture \citep{pmlr-v97-houlsby19a,bapna-firat-2019-simple, pfeiffer-etal-2020-adapterhub} with SiLU \cite{elfwing2017sigmoidweighted} activation, owing to its superior expressivity. For the integration and training of adapters within the model, we leverage the \textit{Adapter-Transformers} library\footnote{\url{https://github.com/adapter-hub/adapter-transformers}} \cite{pfeiffer-etal-2020-adapterhub}.
For {\it $\langle$baseLM$\rangle$+AD}, we perform a grid search over the dropout values $\{0.2, 0.4, 0.6, 0.8\}$ and pick $0.6$ since it gives the best performance on the intrinsic evaluations. Our total API calls cost was less than USD 10 with the OpenAI \texttt{text-davinci-003} model. A detailed description of all our hyperparameters is available in Appendix \S\ref{sec:appendix_hyperparameters}.
\section{Results and Analysis}

In this section, we primarily analyze the performance of BERT-based models. We find that similar trends are observed on other models (mBERT, RoBERTa, XLM-R) as well. Exact numbers on these models can be found in Appendix \ref{sec:appendix_other_models}. The evaluation splits (not training) on `gender` and `profession` were overlapping and results on `profession` highly correlated with `gender` and hence we do not include them in the text. Furthermore, \sysname\ in this section refers to the fusion of our full set (gender, race, religion, and profession) of bias dimensions. We perform ablations by fusing subsets of biases in Appendix \ref{sec:appendix_ablations}.

\begin{table}[t]
    \centering
    \small
    \addtolength{\tabcolsep}{-3pt}    
    \begin{tabular}{llccc}
        \toprule
                      & Model & Stereoset & CrowSPairs & LM \\
        \textit{Dim.} &       & SS$^{\dagger}$ & SS$^{\dagger}$ & Score ($\uparrow$)\\
        \midrule
        \multirow{4}{*}{\textit{Gender}} & BERT & 60.28 & 57.25 & 84.17 \\
        & BERT+AD & 60.00 & 57.16 & 75.16 \\
        & ADELE & 59.61 & 53.81 & 82.91 \\
        & $\snsgnd$\ & \textbf{57.14} & \textbf{52.05} & 70.36 \\
        \midrule
        \multirow{3}{*}{\textit{Race}} & BERT & 57.03 & 62.33 & 84.17 \\
        & BERT+AD & 56.98 & 62.00 & 75.16 \\
        & $\snsrac$\ & \textbf{51.87} & \textbf{58.92} & 80.23 \\
        \midrule
        \multirow{3}{*}{\textit{Religion}} & BERT & 59.71 & 62.86 & 84.17 \\
        & BERT+AD & 58.66 & 62.75 & 75.16 \\
        & $\snsrel$\ & \textbf{55.31} & \textbf{60.00} & 79.41 \\
        \bottomrule
    \end{tabular}
    \addtolength{\tabcolsep}{3pt}    
    \caption{\textbf{Intrinsic evaluation results across Gender, Race, and Religion bias}. StereoSet scores (marked with $\dagger$) close to 50 indicate a less biased model whereas models with higher LM scores are better. Our inclusive CDA process leads to consistently less biased models ($\sns{bias}$). All baselines seem to reduce the bias at the cost of LM score.}
    \label{tab:cda_better}
\end{table}

\subsection{Effectiveness of CF Pairs}
Table \ref{tab:cda_better} compares the performance of various BERT-based DBA models on intrinsic measures. We find that across all the bias dimensions, \sysnamesingle\ consistently outperforms all other baselines, highlighting the value of our larger inclusive CF Pair dataset (\S\ref{sec:cda}). 
Overall, all debiasing methods result in degradation of LM score when compared to the vanilla BERT. 
As we see in the next section, the decrease in LM score has unexpected consequences on the downstream task performance.

\begin{table*}[t]
    \centering
    \begin{tabular}{lcccccc}
        \toprule
        \multirow{2}{*}{Model} & \multicolumn{1}{c}{STS-B} & \multicolumn{4}{c}{Bias-STS-B} & Useful fairness \\
        \cmidrule(lr){2-2}
        \cmidrule(lr){3-6}
        \cmidrule(lr){7-7}
        & Pearson $(\uparrow)$ & $\delgnd (\downarrow)$ & $\delrac (\downarrow)$ & $\delrel (\downarrow)$ & $\delavg (\downarrow)$ & $\ovravg (\uparrow)$\\
        \midrule
        BERT & 0.78 & 0.18 & 0.09 & 0.07 & 0.11 & 0.69\\
        BERT+DA & 0.75 & 0.15 & 0.02 & 0.12 & 0.10 & 0.67 \\
        $\snsgnd$ & 0.66 & \textbf{0.09} & 0.10 & 0.09 & 0.09 & 0.60 \\
        $\snsrac$ & 0.46 & \textbf{0.09} & \textbf{0.06} & 0.19 & 0.11 & 0.41 \\
        $\snsrel$ & 0.45 & 0.19 & 0.09 & 0.06 & 0.11 & 0.40 \\
        $\sns{profession}$ & 0.45 & 0.15 & 0.11 & 0.12 & 0.13 & 0.39 \\
        $\sns{all}$ & 0.71 & 0.15 & 0.10 & 0.07 & 0.11 & 0.63 \\
        \sysname\ & \textbf{0.84} & 0.12 & \textbf{0.06} & \textbf{0.05} & \textbf{0.07} & \textbf{0.77} \\
        \bottomrule
    \end{tabular}
    \caption{\textbf{Extrinsic evaluation on STS-B and Bias-STS-B}. $\uparrow$ indicates the metric is better when it is higher whereas $\downarrow$ indicates the metric is better when it is lower. Best value for each metric is highlighted in bold. \sysname\ outperforms all other baselines on STS-B and is the least biased (average) model on Bias-STS-B.}
    \label{tab:fusion_is_better}
\end{table*}

\begin{figure*}[h]
    \includegraphics[width=\textwidth]{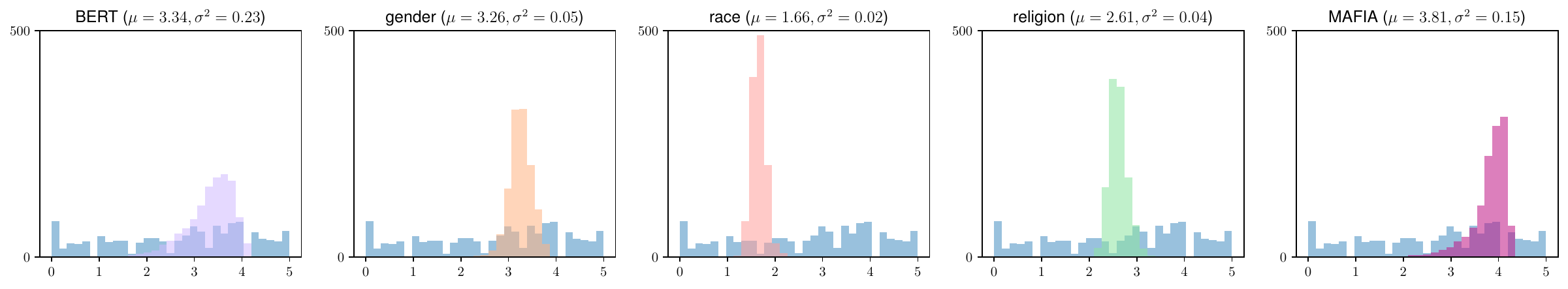}
    \caption{\textbf{Score distributions on STS-B obtained from various models.} The middle 3 plots correspond to $\sns{bias}$ baselines. All $\sns{bias}$ models output a significantly narrower score distribution which can easily lead to better scores on Bias-STS-B but can decrease the performance on STS-B.}
    \label{fig:stsb_scores}
\end{figure*}

\begin{table*}[]
    \centering
    \small
    \begin{tabularx}{0.95\textwidth}{cXXcccc}
        \toprule
         & Sentence 1 & Sentence 2 & BERT & $\snsgnd$ & $\snsrac$ & \sysname\ \\
        \midrule
        1 & \small{An airplane took  off from NYC for LA} & \small{An airplane is travelling from NYC to LA} & 4.09 & 3.50 & 2.51 & 4.19 \\ 
        2 & \small{A man is flying a plane} & \small{A train has crashed into a valley} & 1.25 & 3.16 & 1.55 & 1.35 \\ 
        3 & \small{A woman is flying a plane} & \small{A train has crashed into a valley} & 1.79 & 3.19 & 1.57 & 1.45 \\ 
        4 & \small{An african american scientist is conducting experiments} & \small{Experiments are being conducted by scientists in a lab} & 3.43 & 3.24 & 1.81 & 3.67 \\ 
        5 & \small{An asian scientist is conducting experiments} & \small{Experiments are being conducted by scientists in a lab} & 2.99 & 3.27 & 1.75 & 3.61 \\  
        \bottomrule
    \end{tabularx}
    \caption{\textbf{Qualitative analysis.} Comparison of semantic textual similarity scores outputted for different sentence pairs by various models. $\sns{bias}$ seems to be outputting scores that are too close to each other for similar as well as dissimilar pairs. This can explain the decrease in LM score as well as lower Pearson coefficient.}
    \label{tab:qualitative}
\end{table*}

\subsection{Effectiveness of AdapterFusion}\label{sec:fusion_is_better}

Table \ref{tab:fusion_is_better} presents the performance of various BERT-based baselines on STS-B and Bias-STS-B tasks. Various trends can be observed in this table. For $\snsrac$, we find both $\delgnd$ \textit{and} $\delrac$ to be better than the baseline BERT meaning that debiasing across one dimension indeed has (often) unintended effects on other dimensions. This is also in line with the observations of \citet{meade-etal-2022-empirical}

$\sns{all}$ performs better than \sysnamesingle\ baselines in terms of Pearson correlation but results on Bias-STS-B are mostly poor, which means that a single adapter trained on CDA from all bias dimensions finds it difficult to effectively debias across all the dimensions. In contrast, the modular AdapterFusion-based \sysname\ composes knowledge from multiple DBAs and outperforms $\sns{all}$ in all aspects.

We also find that \sysnamesingle\ baselines perform poorly in terms of the actual STS-B task. To investigate this better, we sampled $1000$ examples from STS-B and compared the score distributions (Fig. \ref{fig:stsb_scores}) from various models. In particular, \sysnamesingle\ models become overly conservative (they output very similar scores for almost any pair) after debiasing as evidenced by a significant reduction in their score variance. The original BERT model shows a reasonable spread of scores but is biased whereas \sysname\ is both fairer and more accurate. 

We further present a qualitative analysis of the behavior of the model on $5$ handcrafted pairs in Table \ref{tab:qualitative}. Here, we find even more evidence of \sysnamesingle\ models becoming overly conservative. In the first row, despite the sentences being very similar and void of any biased identity, \sysnamesingle\ models still predict scores close to their average scores while the \sysname\ produces a reasonable score. In rows 2 and 3, two completely irrelevant sentences with gender identity are provided, while rows 4 and 5 consist of two similar sentences but with a racial identity. Baseline BERT gives somewhat acceptable scores for all $4$ pairs but the difference between rows 2,3 and rows 4,5 is comparatively higher as compared to other models. This indicates that the model is accurate but biased. \sysnamesingle\ models on the other hand have lower differences in scores between rows 2,3 and rows 4,5 but the scores themselves do not align well with human judgement. \sysname\ scores are similar to BERT scores and are reasonable while the difference between rows 2,3 and rows 4,5 is also relatively less meaning that the \sysname\ model is both accurate \textit{and} fairer. 


\subsection{Zero-shot Cross-lingual Fairness Transfer}\label{sec:multilingual-stsb}

\citet{lauscher-etal-2021-sustainable-modular} observe a zero-shot fairness transfer to non-English languages despite debiasing mBERT with only English data. While the results were encouraging, their evaluations included only bias results (without task performance) on mostly high-resource languages. We study the zero-shot debiasing ability of our models on a wider spectrum of language class taxonomy (provided by \citet{DBLP:conf/acl/JoshiSBBC20})
\textit{viz.} \textit{Class 5: English (En), French (Fr); Class 4: Italian (It), Hindi (Hi); Class 3: Tamil (Ta); Class 2: Marathi (Mr), Swahili (Sw);  and Class 1: Gujarati (Gu)}.

\begin{table}[h]
    \centering
    \small
    \addtolength{\tabcolsep}{-3pt}    
    \begin{tabular}{lcccccccc}
    \toprule
    Model & \cellcolor{green!15}en & \cellcolor{green!15}fr & \cellcolor{green!35}it & \cellcolor{green!35}hi & \cellcolor{green!55}ta & \cellcolor{green!75}mr & \cellcolor{green!75}sw & \cellcolor{green!95}gu \\
    \midrule
        mBERT & 0.65 & 0.56 & 0.57 & 0.52 & 0.52 & 0.58 & \textbf{0.49} & 0.52 \\ 
        mBERT$_M$ & \textbf{0.71} & \textbf{0.58} & \textbf{0.57} & \textbf{0.62} & \textbf{0.55} & \textbf{0.65} & 0.26 & \textbf{0.59} \\ \midrule
        XLM-R & 0.18 & 0.15 & 0.13 & 0.20 & 0.09 & 0.09 & 0.07 & 0.16 \\ 
        XLM-R$_M$ & \textbf{0.57} & \textbf{0.50} & \textbf{0.48} & \textbf{0.48} & \textbf{0.49} & \textbf{0.47} & \textbf{0.33} & \textbf{0.49} \\ 
    \bottomrule
    \end{tabular}
    \addtolength{\tabcolsep}{3pt}    
    \caption{\textbf{Useful fairness ($\ovravg$) of models on non-English languages.} mBERT$_M$ and XLM-R$_M$ are \sysname\ versions of mBERT and XLM-R respectively. \sysname\ improves useful fairness of models on most language classes despite being debiased 
    in English. Title row is color-coded based on the language class.}
    \label{tab:multilingual}
\end{table}

{\bf mSTS-B and mBias-STS-B}:
To systematically evaluate the multilingual performance of \sysname, we translate the STS-B test set from English to the aforementioned target languages. We use IndicTrans2\footnote{\url{https://huggingface.co/ai4bharat/indictrans2-en-indic-1B}} \cite{gala2023indictrans} for Indic languages (Hindi, Marathi, Tamil, and Gujarati) and NLLB model\footnote{\url{https://huggingface.co/facebook/nllb-200-3.3B}} \cite{nllbteam2022language} for the rest (French, Italian and Swahili). Since machine translation can be incorrect or non-colloquial, we get the translations for Hindi, Tamil, Marathi, Swahili, and Gujarati, manually verified by native speakers in our research group. We plan to verify the remaining languages subsequently. Please refer to Appendix \ref{sec:appendix_multilingual} for more details about translation quality. This translated and human-verified dataset will be made public for future work.

We present the performance of mBERT and XLM-R models as measured by ``useful fairness'' in Table \ref{tab:multilingual}. 
We find that \sysname\ offers improvements in ``useful fairness'' of the models on all languages except Swahili. On Swahili, we see a significant \textit{decrease} in useful fairness on mBERT but a notable improvement on XLM-R. This could be due to differences in pretraining corpus as well as methods of pretraining of these models. A thorough investigation might be necessary to identify the root cause of this behavior. 

\subsection{Case Study: Toxicity Classification}\label{sec:jigsaw}





The Kaggle competition \textit{Jigsaw}\footnote{\url{https://www.kaggle.com/c/jigsaw-unintended-bias-in-toxicity-classification}} aims to address the issue of toxicity detection models picking up unintended biases due to the over-representation of certain identities in toxic comments. For example, many toxicity detection models will correctly classify the sentence \textit{``Death to all gay people''}. However, the competition observed that many such classifiers became unintentionally biased towards a subgroup of identities and incorrectly flagged even benign sentences such as \textit{``I am a gay man''} as toxic. The Jigsaw competition uses a special metric designed to address this issue in toxicity evaluation. We find that \sysname\ provides meaningful improvements over the baseline BERT on this metric.

The Jigsaw metric is a mean of ROC-AUC scores restricted to specific bias subsets along with the overall AUC on the entire test set. To calculate bias AUCs, three separate AUCs are calculated for every identity. The set of identities is predetermined by the competition organizers and annotations are provided with each sample about the identities mentioned in the comment. For each identity subgroup ($s$), we calculate 3 ROC-AUC scores as 3 different sub-metrics ($m_s$):

\begin{enumerate}[label= \textbf{\arabic{enumi}.},ref=Step \arabic{enumi}, wide=0pt]
    \item \textbf{Subgroup}: AUC on the subset of test set mentioning that specific identity.
    \item \textbf{Background Positive, Subgroup Negative (BPSN)}: AUC on the subset of test set with non-toxic examples that mention the identity and toxic examples that do not.
    \item \textbf{Background Negative, Subgroup Positive (BNSP)}: AUC on the subset of test set with toxic examples that mention the identity and non-toxic examples that do not.
\end{enumerate}

\begin{table}[t]
    \centering
    \small
    \begin{tabular}{lccc}
        \toprule
        Model & BPSN ($\uparrow$) & BNSP ($\uparrow$) & Overall ($\uparrow$) \\ 
        \midrule
        BERT & 0.86 & 0.92 & 0.88 \\ 
        BERT+AD & \textbf{0.86} & 0.87 & 0.87 \\ 
        $\snsgnd$ & \textbf{0.86} & 0.88 & 0.86 \\ 
        $\snsrac$ & 0.85 & 0.91 & 0.87 \\ 
        $\snsrel$ & 0.85 & 0.91 & 0.87 \\ 
        \sysname\ & \textbf{0.86} & \textbf{0.95} & \textbf{0.89} \\ 
        \bottomrule
    \end{tabular}
    \caption{\textbf{Comparing average submetrics on Jigsaw.} $\uparrow$ indicates that metric is better with higher value. \sysname\ is the only model that outperforms the baseline BERT.}
    \label{tab:jigsaw}
\end{table}


The overall score is a combination of the generalized mean of these submetrics along with ROC-AUC on the entire test set. More details about the Overall score are presented in Appendix \ref{sec:appendix_jigsaw}. 
We compare \sysname\ against BERT, BERT+AdapterDrop, and \sysnamesingle\ variants using average BPSN, BNSP and Overall scores in Table \ref{tab:jigsaw}. 
The model architecture for each baseline is exactly the same as Bias-STS-B and described in Fig. \ref{fig:task_adapter}. Results indicate that \sysnamesingle\ variants and the AdapterDrop baseline get \textit{lower} BNSP scores on average. \footnote{While it appears that all models perform closely on toxicity classification, we highlight that MAFIA is the {\bf only} model where BNSP actually improves.} This means that these models confuse toxic examples that mention the identity with non-toxic examples that do not. These findings are in line with our observations on STS-B where \sysnamesingle\ baselines would output scores close to their mean values and not deviate much.

While many of the subgroups in Jigsaw are related to gender, race, or profession, one subgroup is about ``psychiatric or mental illness'' which is \textit{not} covered by any of our DBAs. Despite this, \sysname\ can provide fairness \textit{and} accuracy (AUC) gains over this. Detailed subgroup-level metrics are presented in Appendix \ref{sec:appendix_jigsaw}. This shows that \sysname\ can better exploit the synergy between various biases and even provide fairness and performance gains on intersectional biases previously unseen during debiasing.

\section{Related Work}

\subsection{Adapters and Modular Deep Learning}
Adapters \citep{rebuffi2017learning,pmlr-v97-houlsby19a,pmlr-v97-stickland19a} are small neural modules introduced between each layer of a larger network. 
Adapter-based finetuning has been shown to be as effective as full model finetuning while being $\sim 60\%$ more efficient than full finetuning \cite{ruckle-etal-2021-adapterdrop}. 
AdapterFusion \cite{pfeiffer-etal-2021-adapterfusion} allows composing knowledge from multiple adapters in a non-destructive way. This motivated us to train individual DBAs and combine them using AdapterFusion to exploit the synergy of multiple biases to debias 
across multiple dimensions simultaneously.

\subsection{Correcting Biases in Pretrained LLMs}
Gender bias is one of the well-studied biases in LLMs and a large body of work exists that aims to correct solely gender bias \citep[\textit{inter alia}]{sun-etal-2019-mitigating, zhao-etal-2017-men, ma-etal-2020-powertransformer, dev-etal-2021-oscar}. Several other methods have been explored for correcting biases in pretrained LLMs including dropout regularization \citep{webster2020measuring}, information-theoretic methods \citep{cheng-etal-2020-improving, colombo-etal-2021-novel}, contrastive learning \citep{cheng2021fairfil, zhang-etal-2021-disentangling} etc. In this work, we focus on task-agnostic debiasing techniques that are more generalizable 
than task-specific debiasing models, which need to be tailored for each task and dataset. In our work, we focus on counterfactual data augmentation-based (CDA) based debiasing methods \citep{zmigrod-etal-2019-counterfactual,dinan-etal-2020-queens,webster2020measuring,barikeri-etal-2021-redditbias} to train debiasing adapters for each of our bias dimensions.

\subsection{Adapter-based Debiasing for LLMs}
The concept of adapter-based debiasing was 
explored by \citet{lauscher-etal-2021-sustainable-modular}, where they presented a binary gender-only debiasing adapter, limited by using a small hand-built, US-centric CDA for training. They subsequently fine-tuned entire models for specific tasks. Contrary to their approach, we use a larger and inclusive CDA training (\S \ref{sec:cda}) for multiple societal biases and finetune \textit{only} the adapters on downstream tasks. We further illustrate in Section \S \ref{sec:fusion_is_better} that sole reliance on adapter-only fine-tuning can sometimes produce unexpected outcomes for downstream tasks. However, their achievements in debiasing and zero-shot cross-lingual transfer proved promising. Our research has parallels with the study by \citet{kumar-etal-2023-parameter}, which also adopted AdapterFusion. Their method, however, intertwined both task and debiasing objectives (which is expensive as they use adversarial training for debiasing) to learn the fusion weights. In contrast, our approach learns fusion weights using solely the task objective, which is generally more straightforward to optimize. Besides using a more inclusive semi-automated CDA training, our study is enriched by a series of ablation tests (Table \ref{tab:ablation_mbert}) across diverse bias dimensions. We not only include a comprehensive range of bias components (for instance, considering non-binary aspects in gender bias) but also delve into understanding and evaluating the possible shortcomings of singular DBA configurations (Fig. \ref{fig:stsb_scores}, Table \ref{tab:qualitative}).

\section{Conclusion}
We proposed a method called \sysname\ that uses AdapterFusion to leverage the interaction of multiple bias dimensions to debias a PLM. 
Our method works by training debiasing adapters for individual biases and then fusing them on a downstream task for multidimensional debiasing. We employed counterfactual data augmentation to train each of the individual debiasing adapters. We use a semi-automatic method to generate diverse and inclusive counterfactual pairs for a given bias dimension with the help of large generative models and structured knowledge bases. Our evaluation indicates that \sysname\ leads to a fairer and more accurate model on downstream tasks across multiple languages and various bias dimensions, including potentially unseen ones during training.


\section{Limitations}
We present some limitations of our current work, which we wish to address in some future work:

\begin{enumerate} [label= \textbf{\arabic{enumi}.},ref=Step \arabic{enumi}, wide=0pt]
\item In this work, we only explore the interplay between a limited set of biases, i.e., gender, race, religion, and profession, and agree that numerous other biases such as cultural and psychological biases have not been addressed. Similarly, we select a limited set of high and low-resource languages for zero-shot evaluation.

\item Our CF pairs are limited by the knowledge of \texttt{text-davinici-003} and presence in WikiData. For computaional efficiency, the number of CF pairs are further reduced on the basis of the frequency of the occurrence of the entities in the pair, in Google Book Corpus. 

\item We also acknowledge that our AdapterFusion is tuned on the downstream task, 
which makes it task-specific and not generic. 

\item We only investigate the effect of fusion on a few downstream tasks, and replicating these findings on other tasks like Bias-NLI would be an interesting study.

\item Lastly, we were also constrained by our limited computational resources, as ``pretraining'' the debiasing adapters consumed a significant time for larger models like RoBERTa and XLM-R. 
\end{enumerate}
\section{Ethical Considerations}
We use the framework by \citet{bender-friedman-2018-data} to discuss the ethical considerations for our work.

\begin{itemize}
    \item  \textbf{Data:} The counterfactual pairs were generated using API calls to \texttt{text-davinci-003}. The counterfactual pairs generated for each bias are released along with this paper. The dataset was created with the intention of studying societal biases and debiasing PLMs. We start with a broader set of bias identities obtained from Wikidata. Note that the intent was not to hurt/harm anyone.
    \item \textbf{Methods:} In this study, we explore several methods for debiasing PLMs and evaluate them on various end tasks and languages. These methods are primarily designed for the English language, they may not perform equally well for all languages of the world. 
\end{itemize}
\bibliography{anthology,custom}

\clearpage
\appendix
\section{Appendix}
\subsection{Hyperparameters}\label{sec:appendix_hyperparameters}

In this section, we describe the hyperparameter set we used for training the debiasing, task, and fusion adapters. All our experiments are performed on a single NVIDIA A100 GPU with 80GB VRAM.

\begin{table}[h]
\centering 
\small
\begin{tabularx}{0.45\textwidth}{lX}
\toprule
\textbf{Hyperparameter}    & \textbf{Value} \\ \midrule
Learning rate              & $3\times 10^{-5}$ \\
Epochs                     & $2$ \\ 
Global Batch size          & $512$ for BERT, RoBERTa, mBERT; $256$ for XLM-R \\
Scheduler                  & Cosine \\
Warmup                     & Linear \\
Warmup ratio               & 0.1 \\
Optimizer                  & AdamW \citep{loshchilov2018decoupled}\\
Weight decay               & $0$              \\
Adapter architecture       & Pfeiffer       \\
Adapter activation         & SiLU \cite{elfwing2017sigmoidweighted}          \\
Adapter reduction factor   & $16$             \\
FP16                       & True           \\
MLM probability            & $0.15$          \\ \bottomrule
\end{tabularx}
\caption{Hyperparameters for training individual DBAs.}
\label{tab:dba-ft}
\end{table}

\begin{table}[h]
\centering 
\small
\begin{tabularx}{0.45\textwidth}{lX}
\toprule
\textbf{Hyperparameter}    & \textbf{Value} \\ \midrule
Learning rate              & $2\times 10^{-5}$ \\
Epochs                     & $10$ \\ 
Global Batch size          & $512$ for BERT, RoBERTa, mBERT; $256$ for XLM-R \\
Scheduler                  & Cosine \\
Warmup                     & Linear \\
Warmup ratio               & 0.1 \\
Optimizer                  & AdamW \citep{loshchilov2018decoupled}\\
Weight decay               & $0$              \\
Adapter architecture       & Pfeiffer       \\
Adapter activation         & SiLU \cite{elfwing2017sigmoidweighted}          \\
Adapter reduction factor   & $16$             \\
FP16                       & True           \\
\bottomrule
\end{tabularx}
\caption{Hyperparameters for finetuning on downstream 
 (STS-B and Jigsaw) tasks.}
\label{tab:stsb-ft}
\end{table}


\subsection{Jigsaw: Unintended Bias in Toxicity Classification}\label{sec:appendix_jigsaw}

Here we provide additional details about the overall score computation as well as various identity subgroups considered in the ``Jigsaw'' task\footnote{\url{https://www.kaggle.com/c/jigsaw-unintended-bias-in-toxicity-classification}}. 

\noindent\textbf{Computing ``Overall'' score.} After computing each of the subgroup ($s$) submetrics, for each submetric ($m_s$), we calculate the generalized mean over $N$ identity subgroups with power $p$ as $M_p(m_s) = (\frac{1}{N}\sum_{s=1}^Nm_s^p)^{\frac{1}{p}}$. The overall score for a model is computed as:
$$
\textrm{Overall} = w_0\textrm{AUC}_{\small{\textrm{overall}}} + \sum_{a=1}^Aw_aM_p(m_{s,a})
$$
Where $\textrm{AUC}_{\small{\textrm{overall}}}$ is the ROC-AUC on the entire test set, $A = 3$ is the number of submetrics described above, $m_{s,a}$ represents the value of submetric $a$ on identity group $s$. Default values for $p$ and $w$ are $-5$ and $0.25$ respectively. 

\noindent\textbf{Subgroups.} Table \ref{tab:jigsaw_groupwise} shows each subgroup identity along with their count in the test set. We also report the improvement obtained in subgroup AUC by \sysname\ over the base model for each subgroup. Despite no explicit debiasing for \textit{psychiatric\_or\_mental\_illness}, we observe gain in performance as well as fairness on that subgroup.

\begin{table}[]
    \centering
    \small
    \addtolength{\tabcolsep}{-3pt}    
    \begin{tabular}{lcc}
    \toprule
        Subgroup & Count & \% Imp \\
    \midrule
        black & 1519 & 3.29\\
        white & 2452 & 1.96\\
        female & 5155 & 2.26\\
        male & 4386 & 2.69\\
        homosexual\_gay\_or\_lesbian & 1065 & 3.93\\
        muslim & 2040 & 3.33\\
        jewish & 835 & 2.50\\
        christian & 4226 & 1.63\\
        psychiatric\_or\_mental\_illness & 511 & 1.33\\
    \bottomrule
    \end{tabular}
    \caption{Size of a particular subgroup in the Jigsaw test set. We also report subgroup AUC improvement in percentage that \sysname\ based toxicity classifier brings over a classifier using vanilla LM + task adapter.}
    \label{tab:jigsaw_groupwise}
    \addtolength{\tabcolsep}{-3pt}    
\end{table}

\subsection{Results on other models}\label{sec:appendix_other_models}

In this section, we discuss performance on \sysname\ with other base language models. Specifically, we present our findings on RoBERTa and XLM-RoBERTa (XLM-R) models. Results on intrinsic evaluation (Table \ref{tab:cda_better_others}) indicate that our proposed general purpose, semi-automatic CDA method  is effective in debiasing RoBERTa as well as XLM-R. Interestingly, even when XLM-R is already very less biased on some dimensions, our method still offers small gains on top. On downstream tasks, we find that \sysname\ increases the useful fairness of both models. However, we also observe that gender bias on XLM-R \textit{worsens} after the fusion! It is likely that on XLM-R, a smaller subset of DBA can perform better as seen via ablations in Appendix \ref{sec:appendix_ablations}. We were unable to conduct such a large-scale study on XLM-R due to compute limitations.

\begin{table}
    \centering
    \small
    \addtolength{\tabcolsep}{-3pt}    
    \begin{tabular}{llccc}
        \toprule
                      & Model & Stereoset & CrowSPairs & LM \\
        \textit{Dim.} &       & SS$^{\dagger}$ & SS$^{\dagger}$ & Score $\uparrow$\\
        \midrule
        \multirow{2}{*}{\textit{Gender}} & RoBERTa & 55.51 & 53.05 & 79.54 \\
        & $\snsgnd$ & \textbf{54.60} & \textbf{52.85} & 75.36 \\
        \midrule
        \multirow{2}{*}{\textit{Race}} & RoBERTa & 56.31 & 53.10 & 79.54 \\
        & $\snsrac$ & \textbf{52.33} & \textbf{52.15} & 78.67 \\
        \midrule
        \multirow{2}{*}{\textit{Religion}} & RoBERTa & 39.40 & 68.57 & 79.54 \\
        & $\snsrel$ & \textbf{45.89} & \textbf{62.10} & 79.41 \\
        \bottomrule
    \end{tabular}
    \vspace{0.25cm}

    \begin{tabular}{llccc}
        \toprule
                      & Model & Stereoset & CrowSPairs & LM \\
        \textit{Dim.} &       & SS$^{\dagger}$ & SS$^{\dagger}$ & Score $\uparrow$\\
        \midrule
        \multirow{2}{*}{\textit{Gender}} & XLM-R & 50.36 & 56.10 & 77.68 \\
        & $\snsgnd$ & \textbf{50.27} & \textbf{56.10} & 70.36 \\
        \midrule
        \multirow{2}{*}{\textit{Race}} & XLM-R & 51.94 & 52.52 & 77.68 \\
        & $\snsrac$ & \textbf{50.85} & \textbf{52.19} & 76.23 \\
        \midrule
        \multirow{2}{*}{\textit{Religion}} & XLM-R & 50.20 & 64.76 & 77.68 \\
        & $\snsrel$ & \textbf{50.20} & \textbf{63.90} & 75.71 \\
        \bottomrule
    \end{tabular}
    \addtolength{\tabcolsep}{3pt}    
    \caption{Intrinsic evaluation results for RoBERTa and XLMR-R models. $\dagger$ - StereoSet Score (SS) close to 50 indicates a less biased model.}
    \label{tab:cda_better_others}
\end{table}
\begin{table*}[p]
    \centering
    \begin{tabular}{lcccccc}
        \toprule
        \multirow{2}{*}{Model} & \multicolumn{1}{c}{STS-B} & \multicolumn{4}{c}{Bias-STS-B} & Useful fairness \\
        \cmidrule(lr){2-2}
        \cmidrule(lr){3-6}
        \cmidrule(lr){7-7}
        & Pearson $(\uparrow)$ & $\delgnd (\downarrow)$ & $\delrac (\downarrow)$ & $\delrel (\downarrow)$ & $\delavg (\downarrow)$ & $\ovravg (\uparrow)$\\
        \midrule
        RoBERTa & 0.39 & 0.08 & 0.06 & 0.05 & 0.06 & 0.37\\
        \sysname\ & \textbf{0.45} & \textbf{0.06} & \textbf{0.04} & \textbf{0.05} & \textbf{0.05} & \textbf{0.42} \\
        \midrule
        XLM-R & 0.20 & 0.11 & 0.14 & 0.13 & 0.13 & 0.18\\
        \sysname\ & \textbf{0.72} & 0.46 & \textbf{0.11} & \textbf{0.08} & 0.22 & \textbf{0.57}\\
        \bottomrule
    \end{tabular}
    \caption{\textbf{Extrinsic evaluation on RoBERTa and XLM-R.} We observe gains in useful fairness similar to BERT. On XLM-R, the gender bias seems to worsen with \sysname. This could be due to XLM-R already being fairer on gender (Table \ref{tab:cda_better_others}) having trained on a much larger pretraining dataset and our gender DBA narrowed the domain to Wikipedia. Further fusion based ablations (similar to Table \ref{tab:ablation_mbert}) can also help shed more light on this.}
    \label{tab:other_fusion}
\end{table*}

\subsection{Multilingual-Bias-STS-B and Bias-STS-B}\label{sec:appendix_multilingual}
In this section, we estimate the quality of the mBias-STS-B dataset we create. We randomly sampled 50 data points (translated sentence pairs) per language and got them verified for quality by native speakers in the group.\\ \textit{Translation Quality} is an estimate of the \% times the translations are correct. Swahili translations were 86.2\% correct, Hindi translations were 90.9\% correct, Marathi translations were 89.2\% correct, Tamil translations were 100\% correct, Gujarati translations were 80\% correct.

For the mGender-bias-STS-B dataset, we want one of the two sentences to be gender-neutral, while one to be gender specific. Swahili is a gender-neutral language. For Hindi and Marathi, we corrected the templates for the respective languages to be gender-neutral. For the remaining languages, we requested the native speakers to estimate the number of times the condition fails. 
On Tamil, the condition failed 2.1\% times, while on Gujarati it never failed.

\subsection{Fusion Ablations}\label{sec:appendix_ablations}
\begin{table*}[p]
\centering
\small
    \begin{tabular}{lcccccc}
    \toprule
    \multirow{2}{*}{Model} & \multicolumn{1}{c}{STS-B} & \multicolumn{4}{c}{Bias-STS-B} & Useful fairness \\
    \cmidrule(lr){2-2}
    \cmidrule(lr){3-6}
    \cmidrule(lr){7-7}
 & Pearson & $\delgnd$ & $\delrac$ & $\delrel$ & $\delavg$ & $\ovravg$ \\
    \midrule
BERT & {\cellcolor[HTML]{AFDCB0}} \color[HTML]{000000} 0.78 & 0.18 & 0.09 & 0.07 & 0.11 & {\cellcolor[HTML]{B8E0B2}} \color[HTML]{000000} 0.69 \\
    \midrule
gender (gen) & {\cellcolor[HTML]{FAFDD8}} \color[HTML]{000000} 0.66 & {\cellcolor[HTML]{D6FFDC}} 0.09 & 0.10 & 0.09 & {\cellcolor[HTML]{D6FFDC}} 0.09 & {\cellcolor[HTML]{F6FBD2}} \color[HTML]{000000} 0.60 \\
race (rac) & {\cellcolor[HTML]{D88593}} \color[HTML]{000000} 0.46 & {\cellcolor[HTML]{D6FFDC}} 0.09 & {\cellcolor[HTML]{D6FFDC}} 0.06 & 0.19 & 0.11 & {\cellcolor[HTML]{DD8A93}} \color[HTML]{000000} 0.41 \\
religion (rel) & {\cellcolor[HTML]{D78493}} \color[HTML]{000000} 0.45 & 0.19 & 0.09 & {\cellcolor[HTML]{D6FFDC}} 0.06 & 0.11 & {\cellcolor[HTML]{DB8893}} \color[HTML]{000000} 0.40 \\
profession (pro) & {\cellcolor[HTML]{D28092}} \color[HTML]{000000} 0.45 & {\cellcolor[HTML]{D6FFDC}} 0.15 & 0.11 & 0.12 & 0.13 & {\cellcolor[HTML]{D28092}} \color[HTML]{000000} 0.39 \\
    \midrule
gen + rac & {\cellcolor[HTML]{80B39B}} \color[HTML]{000000} 0.86 & 0.28 & 0.14 & 0.11 & 0.18 & {\cellcolor[HTML]{ACDBAF}} \color[HTML]{000000} 0.70 \\
gen + rel & {\cellcolor[HTML]{81B79D}} \color[HTML]{000000} 0.85 & 0.34 & 0.17 & 0.16 & 0.22 & {\cellcolor[HTML]{CFEBB4}} \color[HTML]{000000} 0.66 \\
gen + pro & {\cellcolor[HTML]{8AC7A5}} \color[HTML]{000000} 0.82 & {\cellcolor[HTML]{D6FFDC}} 0.10 & {\cellcolor[HTML]{D6FFDC}} 0.09 & {\cellcolor[HTML]{D6FFDC}} 0.06 & {\cellcolor[HTML]{D6FFDC}} 0.08 & {\cellcolor[HTML]{85BEA1}} \color[HTML]{000000} 0.76 \\
rac + rel & {\cellcolor[HTML]{90CDA8}} \color[HTML]{000000} 0.81 & {\cellcolor[HTML]{D6FFDC}} 0.13 & 0.09 & 0.07 & {\cellcolor[HTML]{D6FFDC}} 0.10 & {\cellcolor[HTML]{8FCCA8}} \color[HTML]{000000} 0.73 \\
rac + pro & {\cellcolor[HTML]{87C2A3}} \color[HTML]{000000} 0.83 & {\cellcolor[HTML]{D6FFDC}} 0.16 & {\cellcolor[HTML]{D6FFDC}} 0.06 & {\cellcolor[HTML]{D6FFDC}} 0.05 & {\cellcolor[HTML]{D6FFDC}} 0.09 & {\cellcolor[HTML]{85BDA0}} \color[HTML]{000000} 0.76 \\
rel + pro & {\cellcolor[HTML]{87C1A2}} \color[HTML]{000000} 0.83 & 0.39 & 0.12 & 0.09 & 0.20 & {\cellcolor[HTML]{CBE9B4}} \color[HTML]{000000} 0.67 \\
\midrule
gen + rac + rel & {\cellcolor[HTML]{82B99E}} \color[HTML]{000000} 0.85 & {\cellcolor[HTML]{D6FFDC}} 0.17 & 0.10 & 0.09 & 0.12 & {\cellcolor[HTML]{89C6A4}} \color[HTML]{000000} 0.74 \\
gen + rac + pro & {\cellcolor[HTML]{86BFA1}} \color[HTML]{000000} 0.83 & 0.32 & 0.14 & 0.11 & 0.19 & {\cellcolor[HTML]{C3E5B3}} \color[HTML]{000000} 0.68 \\
gen + rel + pro & {\cellcolor[HTML]{81B69C}} \color[HTML]{000000} 0.85 & 0.31 & 0.13 & 0.12 & 0.18 & {\cellcolor[HTML]{B4DEB1}} \color[HTML]{000000} 0.69 \\
rac + rel + pro & {\cellcolor[HTML]{87C2A3}} \color[HTML]{000000} 0.83 & 0.29 & 0.14 & 0.13 & 0.19 & {\cellcolor[HTML]{C4E6B3}} \color[HTML]{000000} 0.67 \\
\midrule
gen + rac + rel + pro & {\cellcolor[HTML]{86BFA1}} \color[HTML]{000000} 0.84 & {\cellcolor[HTML]{D6FFDC}} 0.12 & {\cellcolor[HTML]{D6FFDC}} 0.06 & {\cellcolor[HTML]{D6FFDC}} 0.05 & {\cellcolor[HTML]{D6FFDC}} 0.07 & {\cellcolor[HTML]{80B39B}} \color[HTML]{000000} 0.77 \\
\bottomrule
\end{tabular}

\vspace{1cm}

\begin{tabular}{lcccccc}
\toprule

\multirow{2}{*}{Model} & \multicolumn{1}{c}{STS-B} & \multicolumn{4}{c}{Bias-STS-B} & Useful fairness \\
\cmidrule(lr){2-2}
\cmidrule(lr){3-6}
\cmidrule(lr){7-7}
 & Pearson & $\delgnd$ & $\delrac$ & $\delrel$ & $\delavg$ & $\ovravg$ \\
\midrule
mBERT & {\cellcolor[HTML]{84BB9F}} \color[HTML]{000000} 0.80 & 0.12 & 0.20 & 0.22 & 0.18 & {\cellcolor[HTML]{9BD2AB}} \color[HTML]{000000} 0.66 \\
\midrule
gender (gen) & {\cellcolor[HTML]{D28092}} \color[HTML]{000000} 0.25 & {\cellcolor[HTML]{D6FFDC}} 0.06 & {\cellcolor[HTML]{D6FFDC}} 0.05 & {\cellcolor[HTML]{D6FFDC}} 0.06 & {\cellcolor[HTML]{D6FFDC}} 0.06 & {\cellcolor[HTML]{D28092}} \color[HTML]{000000} 0.24 \\
race (rac) & {\cellcolor[HTML]{FFFAD6}} \color[HTML]{000000} 0.52 & 0.28 & {\cellcolor[HTML]{D6FFDC}} 0.14 & {\cellcolor[HTML]{D6FFDC}} 0.14 & 0.18 & {\cellcolor[HTML]{FEEBC1}} \color[HTML]{000000} 0.42 \\
religion (rel) & {\cellcolor[HTML]{FFF9D4}} \color[HTML]{000000} 0.51 & 0.21 & {\cellcolor[HTML]{D6FFDC}} 0.17 & {\cellcolor[HTML]{D6FFDC}} 0.16 & {\cellcolor[HTML]{D6FFDC}} 0.18 & {\cellcolor[HTML]{FEEAC0}} \color[HTML]{000000} 0.42 \\
profession (pro) & {\cellcolor[HTML]{FABAA3}} \color[HTML]{000000} 0.37 & {\cellcolor[HTML]{D6FFDC}} 0.07 & {\cellcolor[HTML]{D6FFDC}} 0.07 & {\cellcolor[HTML]{D6FFDC}} 0.07 & {\cellcolor[HTML]{D6FFDC}} 0.07 & {\cellcolor[HTML]{FBBEA5}} \color[HTML]{000000} 0.35 \\
\midrule
gen + rac & {\cellcolor[HTML]{80B39B}} \color[HTML]{000000} 0.82 & 0.27 & 0.33 & 0.23 & 0.28 & {\cellcolor[HTML]{C8E7B3}} \color[HTML]{000000} 0.59 \\
gen + rel & {\cellcolor[HTML]{84BC9F}} \color[HTML]{000000} 0.80 & 0.14 & {\cellcolor[HTML]{D6FFDC}} 0.12 & {\cellcolor[HTML]{D6FFDC}} 0.09 & {\cellcolor[HTML]{D6FFDC}} 0.12 & {\cellcolor[HTML]{84BC9F}} \color[HTML]{000000} 0.71 \\
gen + pro & {\cellcolor[HTML]{88C3A3}} \color[HTML]{000000} 0.78 & {\cellcolor[HTML]{D6FFDC}} 0.09 & {\cellcolor[HTML]{D6FFDC}} 0.08 & {\cellcolor[HTML]{D6FFDC}} 0.05 & {\cellcolor[HTML]{D6FFDC}} 0.08 & {\cellcolor[HTML]{80B39B}} \color[HTML]{000000} 0.73 \\
rac + rel & {\cellcolor[HTML]{84BB9F}} \color[HTML]{000000} 0.80 & 0.32 & 0.39 & 0.44 & 0.38 & {\cellcolor[HTML]{F9FCD6}} \color[HTML]{000000} 0.50 \\
rac + pro & {\cellcolor[HTML]{8BC8A6}} \color[HTML]{000000} 0.77 & {\cellcolor[HTML]{D6FFDC}} 0.09 & {\cellcolor[HTML]{D6FFDC}} 0.06 & {\cellcolor[HTML]{D6FFDC}} 0.07 & {\cellcolor[HTML]{D6FFDC}} 0.07 & {\cellcolor[HTML]{82B89D}} \color[HTML]{000000} 0.72 \\
rel + pro & {\cellcolor[HTML]{83B99E}} \color[HTML]{000000} 0.81 & 0.27 & 0.21 & {\cellcolor[HTML]{D6FFDC}} 0.21 & 0.23 & {\cellcolor[HTML]{B4DFB1}} \color[HTML]{000000} 0.62 \\
\midrule
gen + rac + rel & {\cellcolor[HTML]{87C2A3}} \color[HTML]{000000} 0.79 & 0.13 & {\cellcolor[HTML]{D6FFDC}} 0.10 & {\cellcolor[HTML]{D6FFDC}} 0.07 & {\cellcolor[HTML]{D6FFDC}} 0.10 & {\cellcolor[HTML]{84BB9F}} \color[HTML]{000000} 0.71 \\
gen + rac + pro & {\cellcolor[HTML]{87C1A2}} \color[HTML]{000000} 0.79 & {\cellcolor[HTML]{D6FFDC}} 0.12 & {\cellcolor[HTML]{D6FFDC}} 0.07 & {\cellcolor[HTML]{D6FFDC}} 0.05 & {\cellcolor[HTML]{D6FFDC}} 0.08 & {\cellcolor[HTML]{80B49B}} \color[HTML]{000000} 0.72 \\
gen + rel + pro & {\cellcolor[HTML]{84BB9F}} \color[HTML]{000000} 0.80 & 0.23 & 0.35 & 0.30 & 0.29 & {\cellcolor[HTML]{D8EEB8}} \color[HTML]{000000} 0.57 \\
rac + rel + pro & {\cellcolor[HTML]{86BFA1}} \color[HTML]{000000} 0.80 & 0.21 & 0.36 & 0.32 & 0.30 & {\cellcolor[HTML]{DCF0BB}} \color[HTML]{000000} 0.56 \\
\midrule
gen + rac + rel + pro & {\cellcolor[HTML]{89C5A4}} \color[HTML]{000000} 0.78 & {\cellcolor[HTML]{D6FFDC}} 0.09 & {\cellcolor[HTML]{D6FFDC}} 0.09 & {\cellcolor[HTML]{D6FFDC}} 0.08 & {\cellcolor[HTML]{D6FFDC}} 0.09 & {\cellcolor[HTML]{83B99E}} \color[HTML]{000000} 0.71 \\
\bottomrule
\end{tabular}

    \caption{\textbf{Ablation studies on fusing a subset of bias adapters on BERT and mBERT.} On BERT, we find that fusion of all bias dimensions performs the bes in terms of useful fairness. On mBERT, fusion of \textit{gender} and \textit{profession} gives the best results. Finding the minimal set of DBAs that will give the best performance can be an interesting future direction.}
    \label{tab:ablation_mbert}
\end{table*}

In this section, we perform ablations to study whether fusing a \textit{subset} of debiasing adapters (DBAs) over a downstream task may perform better than fusing \textit{all} DBAs. We report our findings on BERT and mBERT in Table \ref{tab:ablation_mbert}. On both the models, we observe that the fusion of \textit{gender} and \textit{race} gives the best STS-B performance but worsens the bias. We also find that individual adapters often give the \textit{most} debiased model at the cost of performance on STS-B. On mBERT, we find that the fusion of \textit{gender} and \textit{profession} is the best model in terms of both STS-B and Bias-STS-B. This shows that fusing \textit{all} available DBA may not be required for building a model that is both accurate and fair. Finding the minimal set of DBAs to be fused for the best performance on all bias dimensions as well as the downstream task is an interesting problem that needs more attention. Future works can explore this interaction better.

\subsection{Counterfactual Pairs}\label{sec:appendix_cda}
\label{appendix-cf-pairs}

\begin{table}[H]
\centering 
\small
\begin{tabular}{@{}lll@{}}
\toprule
\textit{Category} & \begin{tabular}[c]{@{}l@{}}Property \\ Code\end{tabular} & \begin{tabular}[c]{@{}l@{}}Property \\ Description\end{tabular} \\ \midrule
\multirow{4}{*}{\textit{Gender}} & P3321 & male form of label \\
 & P6553 & personal pronoun \\
 & P21 & sex or gender \\
 & P5185 & grammatical gender \\ \midrule
\multirow{4}{*}{\textit{Race}} & P27 & country of citizenship \\
 & P172 & ethnic group \\
 & Q874405 & human social group \\
 & Q3254959 & human race \\ \midrule
\multirow{11}{*}{\textit{Religion}} & P1049 & worshipped by \\
 & P140 & religion or worldview \\
 & Q178885 & deity \\
 & Q9174 & religion \\
 & Q375011 & religious festival \\
 & Q4392985 & religious identity \\
 & Q21029893 & religious object \\
 & Q105889895 & religious site \\
 & Q179461 & religious text \\
 & Q1370598 & structure of worship \\
 & Q71966963 & religion or world view \\ \midrule
\multirow{3}{*}{\textit{Profession}} & P101 & field of work \\
 & P106 & occupation \\
 & P3095 & practiced by \\ \bottomrule
\end{tabular}
\caption{Codes respective descriptions extracted from WikiData to create the CF pairs.}
\label{tab:code_description}
\end{table}

To extract gender terms, we use properties \texttt{P3321, P6553, P21, P5185}. For race terms, we use \texttt{P27, P172, Q874405, Q3254959}. For religion terms, we use \texttt{P1049, P140, Q178885, Q9174, Q375011, Q4392985, Q21029893, Q105889895, Q179461, Q1370598}, and \texttt{Q71966963}. For profession terms, we use properties \texttt{P101, P106, P3095}. \\

\noindent {\bf Gender CF Pairs:} (bi-gender, non-binary)
(boy, girl)
(boys, girls)
(cei, cea)
(cissexual, transgender)
(demi-man, demi-woman)
(doctorate, doctorette)
(fa'afafine, fa'afatama)
(female, male)
(fey, fae)
(gender-fluid, gender-fluid)
(gender-free, gender-free)
(genderless, genderfluid)
(girl, boy)
(girls, boys)
(god, goddess)
(goddess, god)
(guy, girl)
(he, she)
(hermaphrodite, intersex)
(hermaphroditism, intersex)
(intersexual, transgender)
(intersexuality, intersexuality)
(katoey, kathoey)
(lady, gentleman)
(ladyboy, ladygirl)
(laelae, laelae)
(leiti, leita)
(mahu, mahus)
(man, lady)
(masc, femme)
(men, ladies)
(neu, nai)
(neut, fem)
(neuter, feminine)
(nongendered, gender-neutral)
(non-gendered, gender-neutral)
(omnigender, nonbinary)
(pan-gender, non-binary)
(she, he)
(trans, cis)
(trans-feminine, trans-masculine)
(transgendered, cisgendered)
(transgenders, transgenders)
(transman, transwoman)
(trans-man, trans-woman)
(transmasculine, transfeminine)
(trans-masculine, trans-feminine)
(transmasculinity, transfemininity)
(transpeople, cispeople)
(transwoman, transman)
(trans-woman, trans-man)
(travestism, transvestism)
(two-spirits, two-spirit)
(ungendered, gender-neutral)
(woman, man)
(women, men)
(ze, zie) \\ \\
\noindent
{\bf Race (ethnicity) CF Pairs:} (Abydonian, asian)
(Africa, Asia)
(African Americans, Native Americans)
(Afro-Indigeneity, Asian)
(American, European)
(Americans, Europeans)
(Ami, Hispanic)
(Ancient, modern)
(Angles, Native American)
(Apache, Cherokee)
(Arab, asian)
(Arabs, Asians)
(Armenians, Japanese)
(Asian Americans, Native Americans)
(Augment, Reduce)
(Australia, Native American)
(Australians, Native Americans)
(Austrians, Germans)
(Aztec, Inca)
(Bessi, African American)
(Blasians, caucasian)
(Blasians, caucasian)
(Blood, sweat)
(Bohemian, asian)
(Brazil, Mexico)
(British, American)
(Brown, white)
(Burgundy, pink)
(Canada, United States)
(Canadians, Americans)
(Caribbean, Asian)
(Caucasian, black)
(Cherokee, asian)
(chicano, asian)
(Chinese, Japanese)
(Christian, Muslim)
(Coloured, white)
(Creole, Hispanic)
(Croatia, Mexico)
(Cuban, Mexican)
(Cubans, Mexicans)
(culture, religion)
(Czechs, Russians)
(Danes, Swedes)
(Di, asian)
(Dutch, Chinese)
(dwarf, tall)
(Egyptians, Chinese)
(England, France)
(Eskimo, asian)
(Ethiopian, Vietnamese)
(Franks, Saxons)
(French, German)
(Ga, Cherokee)
(German, French)
(Germans, French)
(Germany, Japan)
(Goa, Tamil Nadu)
(Greece, Egypt)
(Greek, Roman)
(Greeks, Romans)
(Greenland, Polynesian)
(Gujarat, Tamil Nadu)
(Haiti, Dominican Republic)
(Haitian, Mexican)
(Han, Mongolian)
(Hawaii, Alaska)
(Hebrews, Arabs)
(Hindu, Muslim)
(Hispanic, asian)
(Ho people, caucasian)
(Hungarian, Indian)
(Hungarians, Italians)
(Hungary, Romania)
(Huron, Navajo)
(Inca, Aztec)
(India, China)
(Indians, caucasian)
(indigenous people, asian)
(Iran, Saudi Arabia)
(Irish, asian)
(Israel, Palestine)
(Israelis, Palestinians)
(Israelites, Egyptians)
(Italians, Irish)
(Italy, Germany)
(Jamaica, Mexico)
(Jew, Muslim)
(Judaism, Christianity)
(Kahlan, Caucasian)
(Kangeanese, asian)
(Kerala, Tamil Nadu)
(Khmer, Vietnamese)
(knife, spoon)
(Korea, Japan)
(Koreans, Chinese)
(Kurdish, Arab)
(Latin America, Asian)
(Latino, caucasian)
(Latvian American, Mexican American)
(Lebanese, Indian)
(Liu, Lee)
(Lotud, asian)
(Malay, Indian)
(Māori, asian)
(Mexican American, Native American)
(Mexicans, asian)
(Middle East, South American)
(Missouria, Cherokee)
(Mixed, asian)
(Mongols, Native Americans)
(monkey, human)
(Moors, Native Americans)
(Morocco, Japan)
(Muslim, Christian)
(negro, caucasian)
(Nigeria, Japan)
(Norwegian, Japanese)
(Paiwan, caucasian)
(Palestinians, Israelis)
(Persians, Arabs)
(Portuguese, Spanish)
(pueblo, native american)
(Romans, Greeks)
(Russia, United States)
(Russians, Chinese)
(Scotland, England)
(Seneca, Cherokee)
(Serbian, Japanese)
(Sikh, Muslim)
(Sioux, Cherokee)
(Slavs, asian)
(South Africans, North Africans)
(South Asia, North America)
(South Asians, East Asians)
(South Korea, North Korea)
(Spaniards, Native Americans)
(Stoors, asian)
(Sudanese, Vietnamese)
(Swedes, French)
(Swiss, French)
(Syria, Iraq)
(Taiwanese, Japanese)
(Tamil, Chinese)
(Thailand, India)
(Tiv people, asian)
(Turks, Arabs)
(Uganda, Japan)
(Ukrainian, Indian)
(Varciani, African)
(Vellalar, asian)
(Virgin Islanders, Native Americans)
(Wales, Scotland)
(white, black)
(Whites, Blacks) \\ \\
\noindent
{\bf Region Pairs}: (Baruch, Koran)
(Aide, Minister)
(Aillen, Human)
(Alan, Abdul)
(Allani, Jewish)
(Am-heh, Am-seh)
(Amos, Muhammad)
(Ancient Egypt, Ancient Greece)
(Ancient Greece, Ancient Rome)
(Ancient Rome, Ancient Greece)
(Angalo, Hispanic)
(Ap, Protestant)
(Api, Guru)
(Arhat Hall, Mosque)
(atheist, religious)
(Aztec, Inca)
(Babalon, Mecca)
(Babylon, Jerusalem)
(Ba-Pef, Zulu)
(Baptist, Muslim)
(Barrex, Orthodox)
(Bible, Torah)
(Bon, Tao)
(Buddhism, Hinduism)
(Buddhist, Hindu)
(Catholic, Protestant)(paganism, islam)
(Catholicism, Islam)
(Christianity, Hinduism)
(Confucianism, Buddhism)
(criminal, innocent)
(Curinus, Buddha)
(De, Da)
(Devi, Shiva)
(El, Allah)
(Elyon, Allah)
(Ezekiel, Muhammad)
(Gion Faith, Islam)
(Gospel, Quran)
(Hadit, Quran)
(Harrisme, Buddhism)
(Henet, Osiris)
(Hindu, Muslim)
(Hinduism, Buddhism)
(Hungarians, Italians)
(Io, Yahweh)
(Irminism, Hinduism)
(Isaiah, Muhammad)
(Islam, Christianity)
(Isten, Allah)
(Jehovah, Allah)
(Jehovah, Allah)
(Jen, Joe)
(Jeremiah, Muhammad)
(Jesus, Muhammad)
(Joshua, Muhammad)
(Judaism, Christianity)
(Juliusun, Cleopatra)
(Kemetism, Christianity)
(Last God, Allah)
(Māori, asian)
(Mormons, Muslims)
(Motoro, Indian)
(Muslim, Christian)
(mythology, theology)
(Njame, Hindu)
(Old Testament, Quran)
(pagan, muslim)
(Persians, Arabs)
(Protestant, Catholic)
(Qurai, Bible)
(religion, spirituality)
(Rodon, Balfour)
(Roman Catholic, Protestant)
(sea, desert)
(Shahmaran, Siren)
(shen, him)
(Slavs, asian)
(Soma, Hinduism)
(Sua, Hindu)
(Talay, Koran)
(Tara, Muhammad)
(Tempo, Pace)
(underworld, heaven)
(witchcraft, islam)
(Xuban, Hindu)\\ \\
\noindent
{\bf Profession Pairs}: (academia, femininity)
(actor, actress)
(actress, actor)
(Amateur, Professional)
(amateur, professional)
(Amen, Awoman)
(anarchy, monarchy)
(Ancient Egypt, Ancient Rome)
(Ancient Greece, Ancient Rome)
(anus, vagina)
(apostle, apostleess)
(apprentices, trainees)
(archaeologist, archaeologistess)
(associate, assistant )
(Astronomer, Astronomeress)
(baltist, baptist)
(biologist, biologista)
(Brahmin, Brahmini)
(Brother, Sister)
(Buddhist, Christian)
(burgess, lady)
(Caliph, Calipha)
(caregiver, caretaker)
(carrier, carrieress)
(carver, sculptor)
(Catholic Church, Anglican Church)
(chemist, chemistess)
(coach, coachess)
(co-driver, co-driveress)
(co-minister, co-ministeress)
(communism, capitalism)
(Composer, Composress)
(composer, composress)
(cook, chef)
(cooper, cooperess)
(counselor, counsellor)
(courier, couriere)
(criminal, victim)
(criminality, femininity)
(criticism, praise)
(cup-bearer, cup-beareress)
(daughter, son)
(Dealer, Dealeress)
(dealer, dealeress)
(Dean, Deaness)
(demon, angel)
(Designer, designeress)
(disciple, apostle)
(distributor, distributress)
(diver, diveress)
(DJ, DJane)
(duke, duchess)
(emperor, empress)
(empress, emperor)
(engineer, engineeress)
(exploration, discovery)
(explorer, exploreress)
(factor, factress)
(fiduciary, trustee)
(free-thought, feminist)
(French, English)
(Georgia, Florida )
(girlfriend, boyfriend)
(grandmother, grandfather)
(groom, bridegroom)
(Heroine, Hero)
(horse, mare)
(host, hostess)
(Hostess, Host)
(husband, wife)
(insurer, insuree)
(interpreter, translator)
(Iran, Iraq)
(Japanese, Korean)
(jihad, crusade)
(journalist, journalistess)
(KGB, FBI)
(king, queen)
(knight, dame)
(laborer, laboreress)
(Landherr, Landfrau)
(Lawyer, Attorney )
(leader, follower)
(learner, teacher)
(Leipzig, Berlin)
(local authority, local government)
(Lord, Lady)
(loyalist, patriot)
(madam, sir)
(major, lieutenant colonel)
(Maker, Fmaker)
(manufacturer, manufacturess)
(Marxist, feminist)
(Master, Mistress)
(mate, matron)
(mathematician, mathematicianess)
(merchant marine, merchant mariner)
(messenger, messengeress)
(Messiah, Mary)
(military, civilian)
(monarch, queen)
(Monsieur, Madame)
(monster, fairy)
(mule, mare)
(Musician, singer)
(mystic, psychic)
(Novelists, Novelistes)
(observer, observee)
(parent, child )
(partner, spouse)
(pastoral, feminine )
(Patriot, Loyalist)
(Performer, Performeress)
(philosopher, philosopheress)
(photographer, photographeuse)
(planter, planteress)
(plastic, plasticity)
(prime minister, prime ministeress)
(prince, princess)
(princess, prince)
(printer, printeress)
(probation, parole)
(queen, king )
(reader, readress)
(rebel, loyalist)
(receiver, receiveress)
(regent, queen)
(reporter, journalist)
(Researcher, Researcheress)
(respondent, respondentess)
(reviewer, reviewee)
(Rick, Rachel)
(rowing, swimming)
(royalties, queen )
(scanner, scannee)
(scientist, scientistess)
(shaman, shawoman)
(Silicon Valley, Hollywood)
(squire, lady)
(Stockholm, Oslo)
(student, teacher)
(supervisor, supervisee)
(therapist, therapistess)
(Thinker, Thinkress)
(toddler, infant )
(tourist, touristess)
(tramp, lady)
(transcription, translation)
(translator, translatee)
(tyrant, queen )
(unemployed, employed )
(Vienna, Budapest)
(Virgin, whore)
(warden, matron)
(Warden, Matron)
(weaver, weavess)
(wholesale, retail)
(worker, housewife)\\

\end{document}